\RequirePackage{fix-cm}
\RequirePackage{fixltx2e}
\documentclass[twoside,english,aip, jcp,floatfix]{revtex4-1}
\usepackage{mathpazo}

\usepackage[T1]{fontenc}
\usepackage[latin9]{inputenc}
\usepackage[a4paper]{geometry}
\usepackage{xcolor}
\usepackage{babel}
\usepackage{amsmath}
\usepackage{amssymb}
\usepackage{graphicx}
\usepackage[unicode=true,pdfusetitle,
 bookmarks=false,
 breaklinks=true,pdfborder={0 0 0},pdfborderstyle={},backref=false,colorlinks=true]
 {hyperref}
\hypersetup{
 citecolor = blue, linkcolor = blue,  urlcolor  = blue}

\makeatletter

\providecommand{\tabularnewline}{\\}

\usepackage{orcidlink}

\makeatother

\begin{document}

\title{{\Large{}On the good reliability of an interval-based metric to validate
prediction uncertainty for machine learning regression tasks}}

\author{Pascal PERNOT \orcidlink{0000-0001-8586-6222}}

\affiliation{Institut de Chimie Physique, UMR8000 CNRS,~\\
Université Paris-Saclay, 91405 Orsay, France}
\email{pp@caladenia.net}

\begin{abstract}
\noindent {\normalsize{}This short study presents an opportunistic
approach to a (more) reliable validation method for prediction uncertainty
average calibration. Considering that variance-based calibration metrics
(ZMS, NLL, RCE...) are quite sensitive to the presence of heavy tails
in the uncertainty and error distributions, a shift is proposed to
an interval-based metric, the Prediction Interval Coverage Probability
(PICP). It is shown on a large ensemble of molecular properties datasets
that (1) sets of }\emph{\normalsize{}z}{\normalsize{}-scores are well
represented by Student's-$t(\nu)$ distributions, $\nu$ being the
number of degrees of freedom; (2) accurate estimation of 95\,\% prediction
intervals can be obtained by the simple $2\sigma$ rule for $\nu>3$;
and (3) the resulting PICPs are more quickly and reliably tested than
variance-based calibration metrics. Overall, this method enables to
test 20\,\% more datasets than ZMS testing. Conditional calibration
is also assessed using the PICP approach.}\\
\textbf{\textcolor{teal}{\normalsize{}}}\\
\textbf{\textcolor{teal}{\normalsize{}}}\\
\textbf{\textcolor{teal}{\normalsize{}Note}}\textcolor{teal}{\normalsize{}:
the present study is essentially an addendum to Ref.\citep{Pernot2024},
and the reader is invited to consult this reference for details on
the notations and concepts.}{\normalsize\par}
\end{abstract}
\maketitle

\section{Introduction}

A recent study\citep{Pernot2024} showed that the reliability of average
calibration statistics for prediction uncertainties and of their validation
is strongly affected by the shape of the uncertainties ($u_{E}$)
and error ($E$) distributions, and notably by the presence of heavy
tails or outliers. If one considers the ZMS statistic, the mean squares
value of \emph{z}-scores $Z=E/u_{E}$, it was shown that datasets
with $\beta_{GM}(Z^{2})\ge0.8$ cannot be reliably tested for calibration
using the $ZMS=1$ test, where $\beta_{GM}(.)$ is a robust skewness
metric\citep{Groeneveld1984,Bonato2011,Pernot2021}. For a recently
published database of 33 datasets of ML materials properties\citep{Jacobs2024},
this means that only about half of them could reliably be tested for
ZMS calibration. Things are even worse for the RCE statistic {[}$RCE=\left(\sqrt{<u_{E}^{2}>}-\sqrt{<E^{2}>}\right)/\sqrt{<u_{E}^{2}>}${]}
, which is sensitive to both $u_{E}$ and $E$ distributions. This
limits considerably the applicability of variance-based calibration
statistics. 

The present proposition draws on three points that will be detailed
below:
\begin{itemize}
\item the distributions of $Z$ values are very close to scaled Student's-\emph{t},
noted $t_{s}(\nu)$;
\item the $t_{s}(\nu)$-based enlargement factor $k_{95}$ to convert the
standard deviation of $Z$ to the half-range of a 95\,\% probability
interval does not vary strongly with the $\nu$ parameter for $\nu>3$;
\item for large datasets, testing a prediction interval by its coverage
probability is more reliable and less costly than testing a ZMS value,
which requires bootstrapping\citep{Pernot2024}.
\end{itemize}
These observations enable to define a simpler alternative approach
to average calibration validation. Sect.\,\ref{sec:Intervals-based-average-calibrat}
develops and justifies the three central points of the approach. PICP
testing is then applied to Jacobs \emph{et al.}'s datasets and compared
with ZMS testing (Sect.\,\ref{sec:Application}), and extended to
conditional calibration. A brief conclusion is provided next. 

\section{Interval-based average calibration testing\label{sec:Intervals-based-average-calibrat}}

\subsection{PICP and its validation\label{subsec:PICP-and-its}}

In practice, the PICPs are estimated as frequencies over a validation
set\citep{Pernot2022b}
\begin{equation}
PICP_{p}=\frac{1}{M}\sum_{i=1}^{M}\boldsymbol{1}\left(|Z|\le k_{p}\right)\label{eq:PICP-freq}
\end{equation}
where $p$ is a percentage value, $\boldsymbol{1}(x)$ is the \emph{indicator
function} for proposition $x$, taking values 1 when $x$ is true
and 0 when $x$ is false. Confidence intervals $[PICP_{p}^{-},PICP_{p}^{+}]$
on PICP values are derived from properties of the binomial distribution\citep{Pernot2022b}.
The continuity-corrected Wilson method\citep{Newcombe1998} is used
here, following the recommendation of Pernot\citep{Pernot2022a},
Sect.\,D.1. Note that the datasets studied here are large enough
to avoid the potential problems described in this study. 
\begin{figure}[t]
\noindent \begin{centering}
\begin{tabular}{cc}
\includegraphics[width=0.48\textwidth]{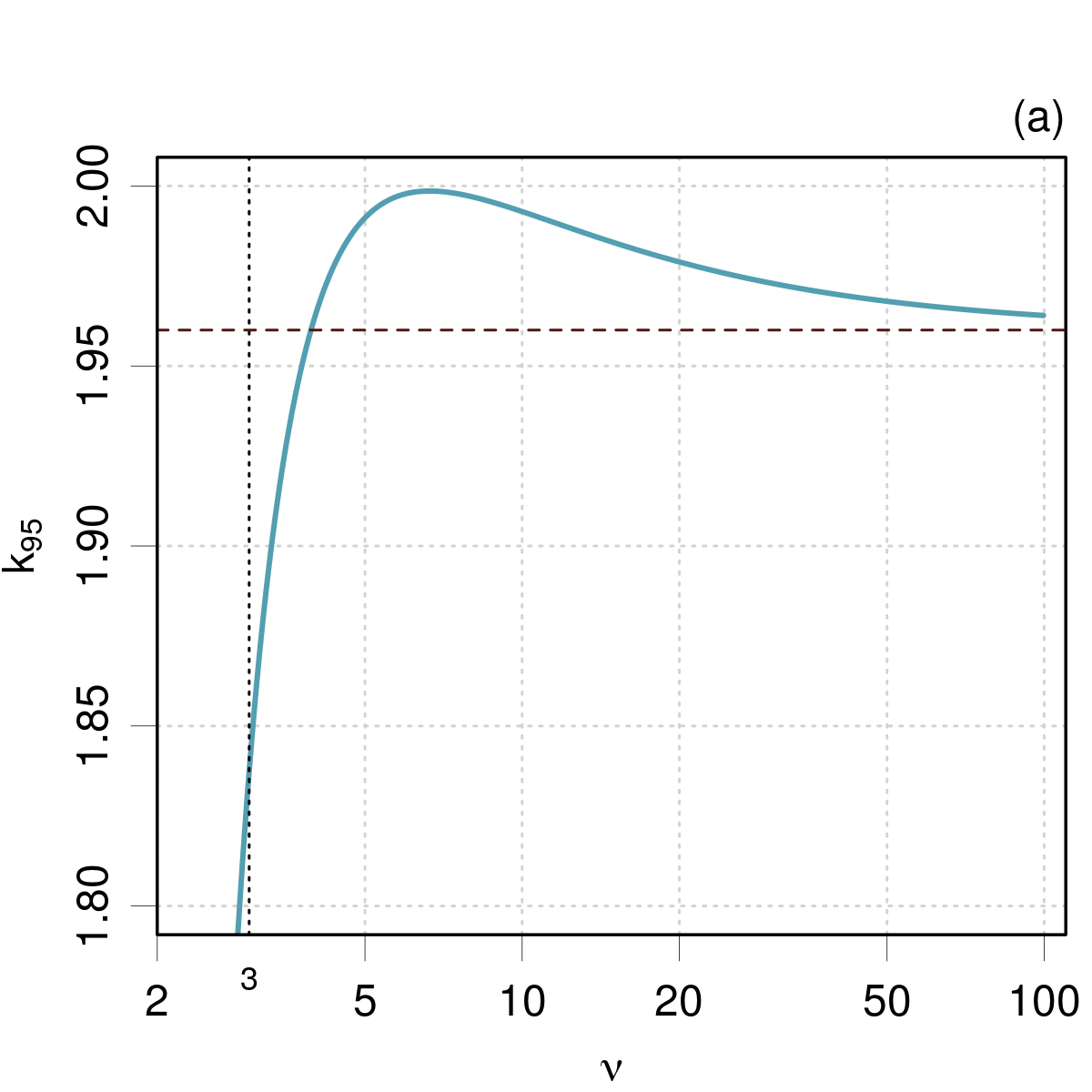} & \includegraphics[width=0.48\textwidth]{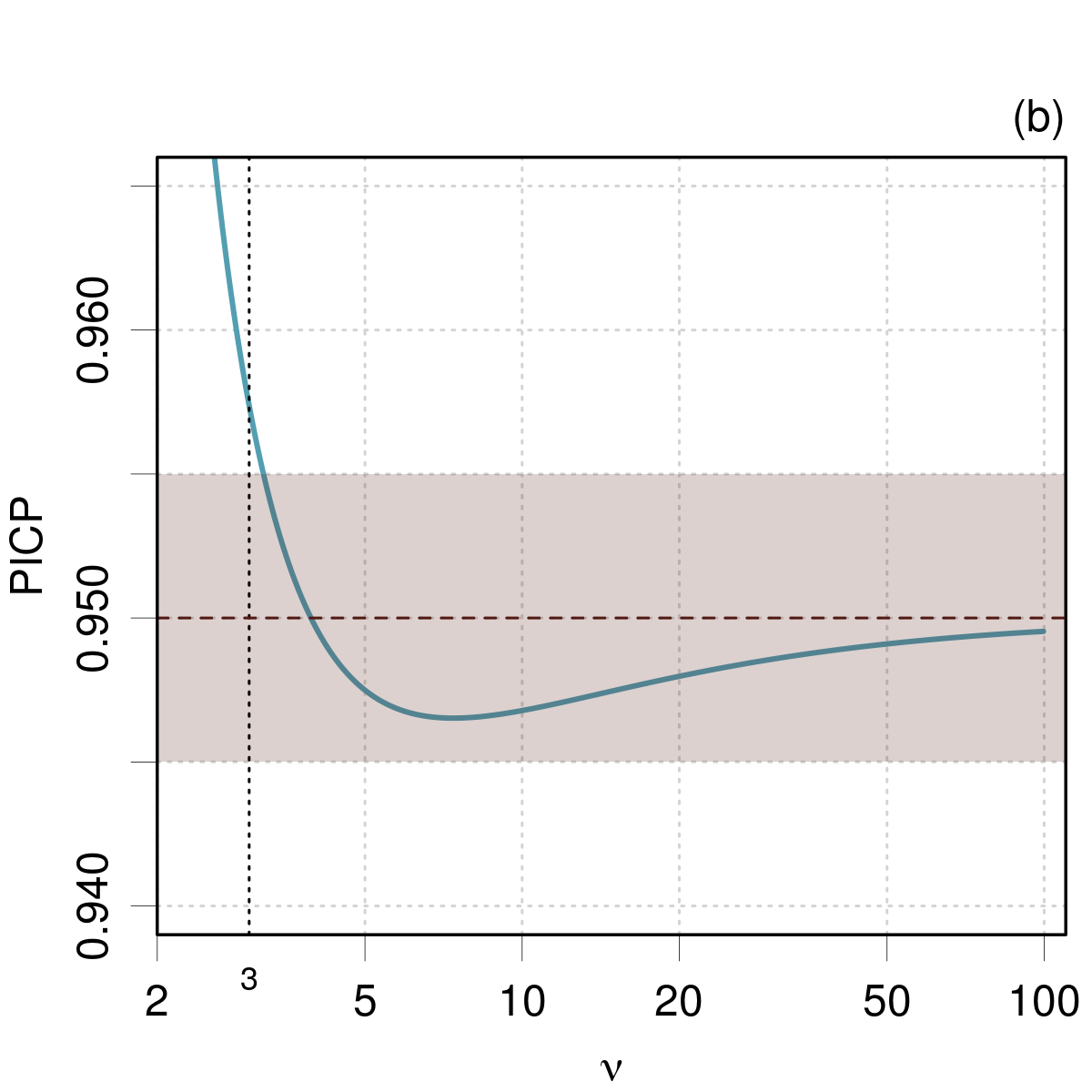}\tabularnewline
 & \includegraphics[width=0.48\textwidth]{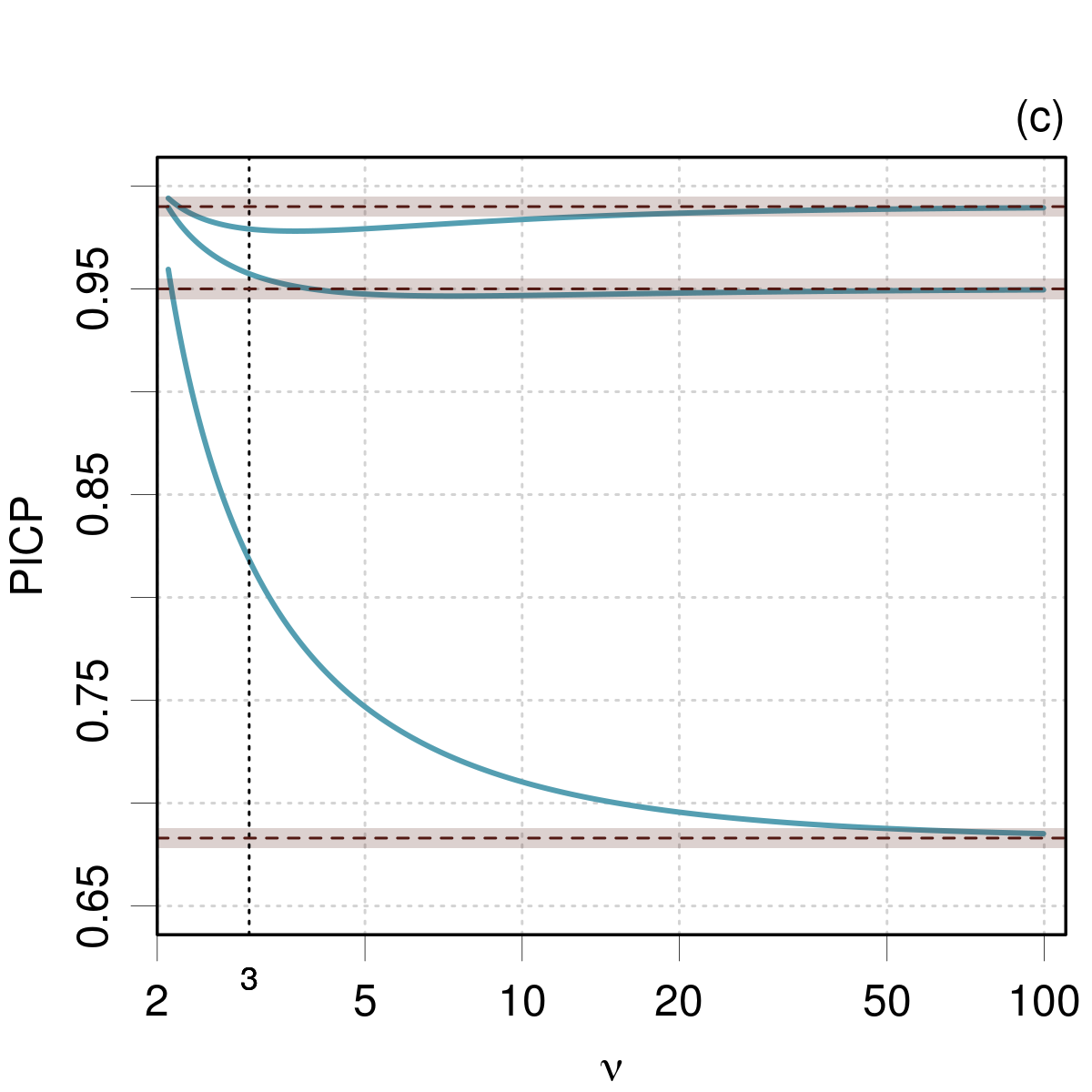}\tabularnewline
\end{tabular}
\par\end{centering}
\caption{\label{fig:PICP}Enlargement factor and coverage probabilities for
a $t_{s}(\nu)$ distribution as a function of $\nu$: (a) $k_{95}$
; (b) coverage probability of a $[-a,a]$ interval for (b) $a=1.96$;
(c) $a=1,\,1.96,\,2.83$. The grayed area depicts a 0.005 deviation
around the asymptotic value.}
\end{figure}

Formally, testing a PICP value for a $p$\,\% prediction interval
is based on checking that the target probability belongs to the PICP
confidence interval, i.e.
\begin{equation}
p/100\in[PICP_{p}^{-},PICP_{p}^{+}]\label{eq:PICPval}
\end{equation}

\subsection{Coverage of $2\sigma$ intervals for $t_{s}(\nu)$ as a function
of $\nu$\label{subsec:Coverage-of-}}

The enlargement factor $k_{95}$ to convert $u_{E}$ to $U_{95}=k_{95}u_{E}$
(the half range of a 95\,\% confidence/prediction interval) for \emph{unit-variance}
$t_{s}(\nu)$ stays very close to the normal asymptotic value (1.96)
for $\nu\ge3$. One can see on Fig.\,\ref{fig:PICP}(a) that it is
non-monotonous with $\nu$ and varies between 1.85 and 2, with a maximum
at about $\nu=6.7$. The $k_{95}=1.9$ value is reached at $\nu\simeq3.3$.
In such conditions, using $k_{95}=1.96$ instead of the exact value
represents at most a 6\,\% error on the enlargement factor. Using
a fixed enlargement factor is important, as it avoids altogether to
fit the $Z$ sample by a $t_{s}(\nu)$ distribution. 

Let us consider the reverse problem, i.e. how the expected coverage
probability varies with $\nu$ when using a fixed value $k_{95}=1.96$.
One sees on Fig.\,\ref{fig:PICP}(b) that the deviation from the
0.95 target is at most 0.005 for $\nu>3$, which is very small when
compared to the uncertainty on empirical PICP values expected for
moderately sized samples\citep{Pernot2022a}. Note that this observation
is not generalizable to other intervals, such as $1\sigma$ or $3\sigma$
, as can be seen on Fig.\,\ref{fig:PICP}(c), where the PICP values
deviate more strongly from their target than in the $2\sigma$ case
(in all rigor, one should write $1.96\sigma)$. 

In order to relax slightly the validation criterion to conform with
a maximal deviation from the theoretical value of 0.005, the following
test is used in practice instead of Eq.\,\ref{eq:PICPval} 
\begin{equation}
0.945\le PICP_{95}^{+}\,\&\,PICP_{95}^{-}\le0.955
\end{equation}

\subsection{Simulations\label{subsec:Simulations}}

\begin{figure}[t]
\noindent \begin{centering}
\includegraphics[width=0.95\textwidth]{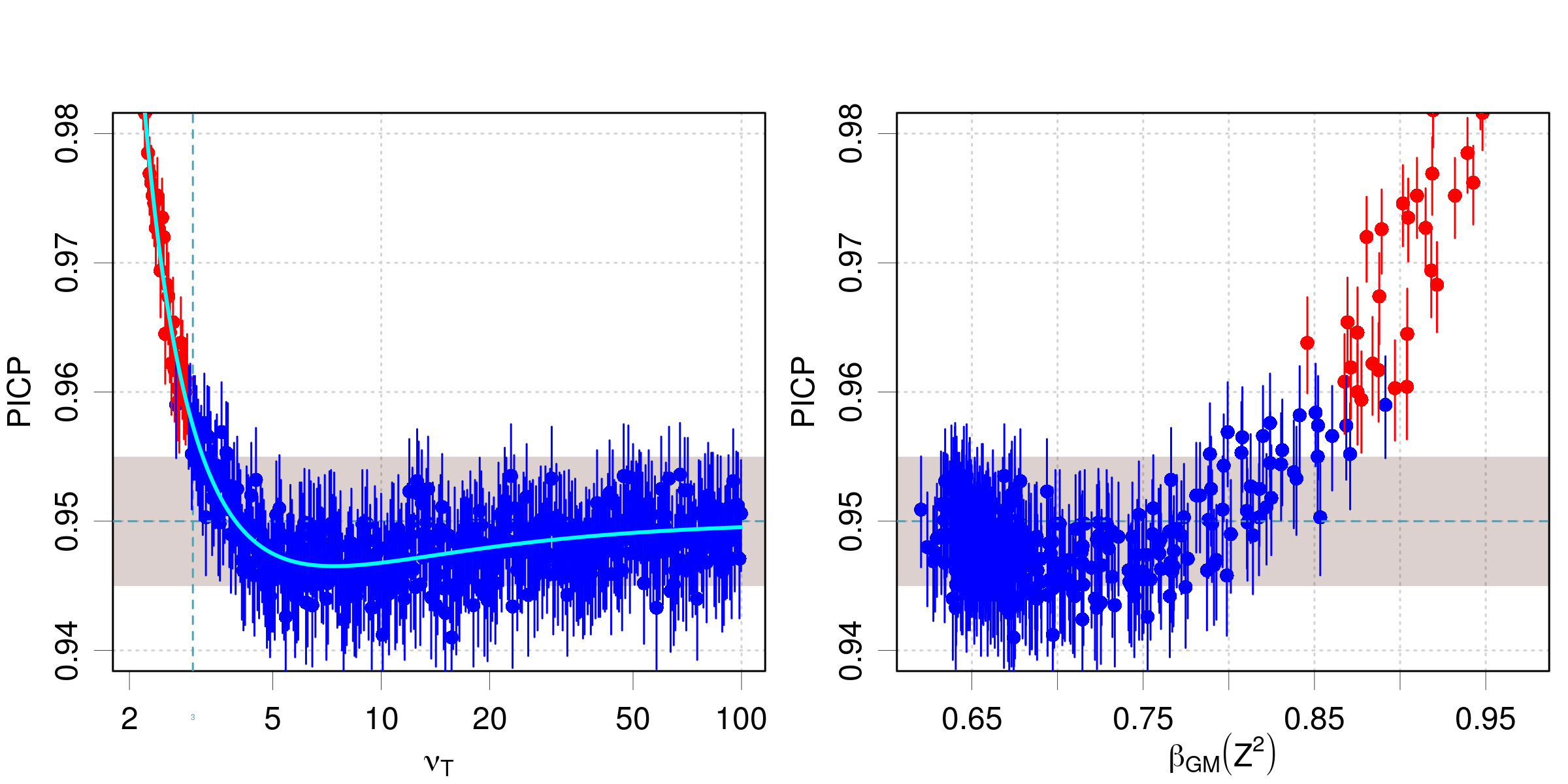}
\par\end{centering}
\caption{\label{fig:simul}PICP$_{95}$ values for $t(\nu)$ samples: (left)
effective coverage as a function of $\nu$; (right) same data as a
function of the $Z^{2}$ skewness. The 95\% confidence intervals on
the PICP values are displayed as error bars. The cyan curve is the
theoretical curve, as seen in Fig.\,\ref{fig:PICP}(b). The gray
area depicts the validity interval. The red points depict invalidated
intervals that do not overlap the gray area. }
\end{figure}

PICP$_{95}$ has been estimated for a series of $Z$ samples generated
from $t_{s}(\nu)$\emph{ }distributions with degrees of freedom varying
between 2 and 100. The samples contain $M=10000$ points. The results
are reported in Fig.\,\ref{fig:simul}(left). One sees that invalid
intervals are essentially obtained for $\nu\le3$, in agreement with
Sect.\,\ref{subsec:Coverage-of-}. When converted to $Z^{2}$ skewness,
this corresponds to $\beta_{GM}(Z^{2})\ge0.85$ {[}Fig.\,\ref{fig:simul}(left){]}.\textcolor{orange}{{}
}These thresholds expand significantly the reliability range for testing
when compared to the ZMS values ($\nu\ge6$ or $\beta_{GM}(Z^{2})\le0.8$)\citep{Pernot2024}.

Note that for ZMS, the threshold for $\beta_{GM}(Z^{2})$ was estimated
from the loss of reliability in the estimation of confidence intervals
by bootstrapping\citep{Pernot2024}, while for PICP, it is based on
an error limit for using a fixed enlargement factor ($k_{95}=1.96$). 

\section{Application\label{sec:Application}}

Jacobs \emph{et al.}\citep{Jacobs2024} published an ensemble of 33
datasets of ML materials properties, with predictions by random forest
models. The prediction uncertainties in these datasets have been calibrated
\emph{pos-hoc,} by polynomial transformation. 

After checking the Student's distribution hypothesis for \emph{z}-scores,
the PICP analysis is performed and compared to the results for ZMS
reported in the Appendix of Ref.\citep{Pernot2024}.\textcolor{orange}{{} }

\subsection{Shape of $Z$ distributions}

Assuming that $Z$ has a unit-variance Student's-$t$ distribution
with $\nu$ degrees of freedom ($Z\sim t_{s}(\nu)$), $Z^{2}$ has
a unit-variance Fisher-Snedecor distribution\citep{Evans2000} with
degrees of freedom $1$ and $\nu$ ($Z^{2}\sim F_{s}(1,\nu)$)\citep{Pernot2024}.
To assess this distribution hypothesis, the $Z^{2}$ datasets have
been fitted by a scaled $F$ distribution, and the results are reported
in the Appendix\,\ref{sec:Distributions-of} (Figs.\,\ref{fig:distZ2}-\ref{fig:distZ2-5}).
The fits are done by maximum goodness-of-fit estimation using the
Kolmogorov-Smirnov distance \citep{Delignette2015}. For each dataset,
the quality of the fit is estimated by visual comparison of the histogram
of $Z^{2}$ values with the best fit density function. 

Indeed, the fits are very good for all sets, except for Set 1 and
13, which are furthermore not testable (see below). Note that Sets
1 and 13 have also been pointed out\citep{Pernot2024} for having
a large fraction of null \emph{z}-scores, which is likely to affect
the fitting. The shape parameters (reported in the title of each plot)
cover a wide range, from 1 to 1743, with a number of very small values
(below 6)\citep{Pernot2024} revealing heavy tails. Large values indicate
quasi-normal $Z$ distribution. 

It has to be noted that the best-fit values of the shape parameter
$\nu$ might be sensitive to the choice of fitting method and to the
adequacy of the chosen distribution. The present values are only indicative
and a complex uncertainty analysis involving model inadequacy, parametric
and statistical uncertainties would be necessary to derive reliable
estimates\citep{Pernot2017}. This is why $\nu$ is not used in practice
as a threshold for the selection of testable datasets. Nevertheless,
the overall quality of the fits confirms the pertinence of the Student's
distribution hypothesis for the studied\emph{ z}-scores.

\begin{figure}[t]
\noindent \begin{centering}
\includegraphics[width=0.95\textwidth]{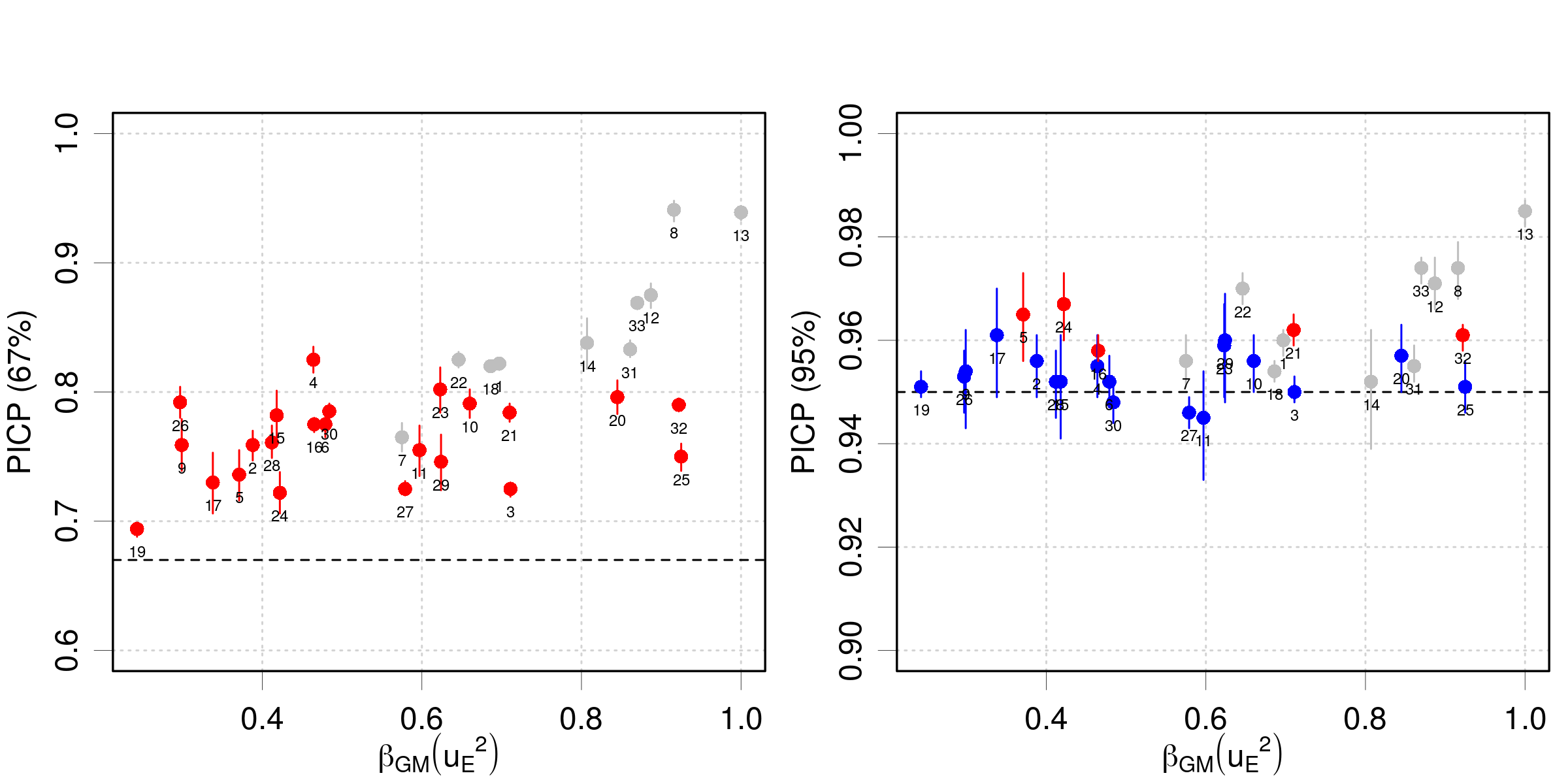}
\par\end{centering}
\caption{\label{fig:LCP}PICP analysis at the $1\sigma$ (left) and $2\sigma$
(right) levels. The 95\% confidence intervals on the PICP values are
displayed as error bars. The points are color-coded into three classes:
(1) gray for sets with $\beta_{GM}(Z^{2})\ge0.85$; (2) blue for calibrated
sets; and (red) for uncalibrated sets.}
\end{figure}

\subsection{Untestable datasets}

It was stated in Sect.\,\ref{subsec:Simulations} that datasets with
$\beta_{GM}(Z^{2})\ge0.85$ cannot be reliably tested by the proposed
PICP metric. This concerns 10 sets: 1, 7, 8, 12, 13, 14, 18, 22, 31
and 33, a subset of the 16 sets with $\beta_{GM}(Z^{2})\ge0.80$ that
would be unsuitable for testing by their ZMS value. Using PICP enables
thus to test about 20\,\% more sets than ZMS. 

\subsection{PICP analysis}

The PICP at the $1\sigma$ and $2\sigma$ levels for the 33 datasets
are reported in Fig.\,\ref{fig:LCP}, where the points are sorted
according to the skewness of the squared uncertainties distribution.
One sees that none of the PICP$_{67}$ values are compatible with
the 67\,\% target. The values are all overestimated, with a positive
trend according to $\beta_{GM}(u_{E}^{2})$. The untestable sets occur
for $\beta_{GM}(u_{E}^{2})\ge0.6$. 

At the PICP$_{95}$ level, a large portion of the testable sets (18/23)
is validated. It is remarkable that the rejected sets (5, 16, 21,
24 and 32) have PICP values in slight excess only, not exceeding 0.97. 

A contingency table is used for the comparison with the results of
ZMS validation (Table\,\ref{tab:Contingency}), based on the classification
of the datasets into three classes (valid, invalid and untestable).
If one considers the 17 datasets for which the validity comparison
can be made (16 are excluded by ZMS), the PICP and ZMS metrics agree
for 13 of them, but conflict for 4 (Sets 6, 16, 28 and 32). From the
6 datasets that were deemed untestable by ZMS and testable by PICP,
5 are validated. So globally, PICP validates 18 sets and invalidates
5, where ZMS validated 13 and invalidated 4. 
\begin{table}[t]
\noindent \centering{}%
\begin{tabular}{l|l|r@{\extracolsep{0pt}.}lr@{\extracolsep{0pt}.}lr@{\extracolsep{0pt}.}l|r@{\extracolsep{0pt}.}l|}
\multicolumn{1}{l}{} & \multicolumn{1}{l}{} & \multicolumn{8}{c}{ZMS}\tabularnewline[\doublerulesep]
\cline{3-10} 
\multicolumn{1}{l}{} &  & \multicolumn{2}{c}{~valid~ } & \multicolumn{2}{c}{~invalid~ } & \multicolumn{2}{c|}{~untestable~ } & \multicolumn{2}{c|}{~Sum~ }\tabularnewline
\cline{2-10} 
 & ~valid  & \multicolumn{2}{c}{11 } & \multicolumn{2}{c}{2 } & \multicolumn{2}{c|}{5 } & \multicolumn{2}{c|}{18}\tabularnewline
PICP~ & ~invalid & \multicolumn{2}{c}{2 } & \multicolumn{2}{c}{2 } & \multicolumn{2}{c|}{1 } & \multicolumn{2}{c|}{5}\tabularnewline
 & ~untestable~  & \multicolumn{2}{c}{0 } & \multicolumn{2}{c}{0 } & \multicolumn{2}{c|}{10 } & \multicolumn{2}{c|}{10}\tabularnewline
\cline{2-10} 
 & ~Sum  & \multicolumn{2}{c}{13 } & \multicolumn{2}{c}{4 } & \multicolumn{2}{c|}{16 } & \multicolumn{2}{c|}{33}\tabularnewline
\cline{2-10} 
\end{tabular}\caption{\label{tab:Contingency}Contingency table for the validation by ZMS
(rows) and PICP (columns) metrics.}
\end{table}

\subsection{LCP analysis}

\begin{figure}[t]
\noindent \begin{centering}
\includegraphics[width=0.9\textwidth]{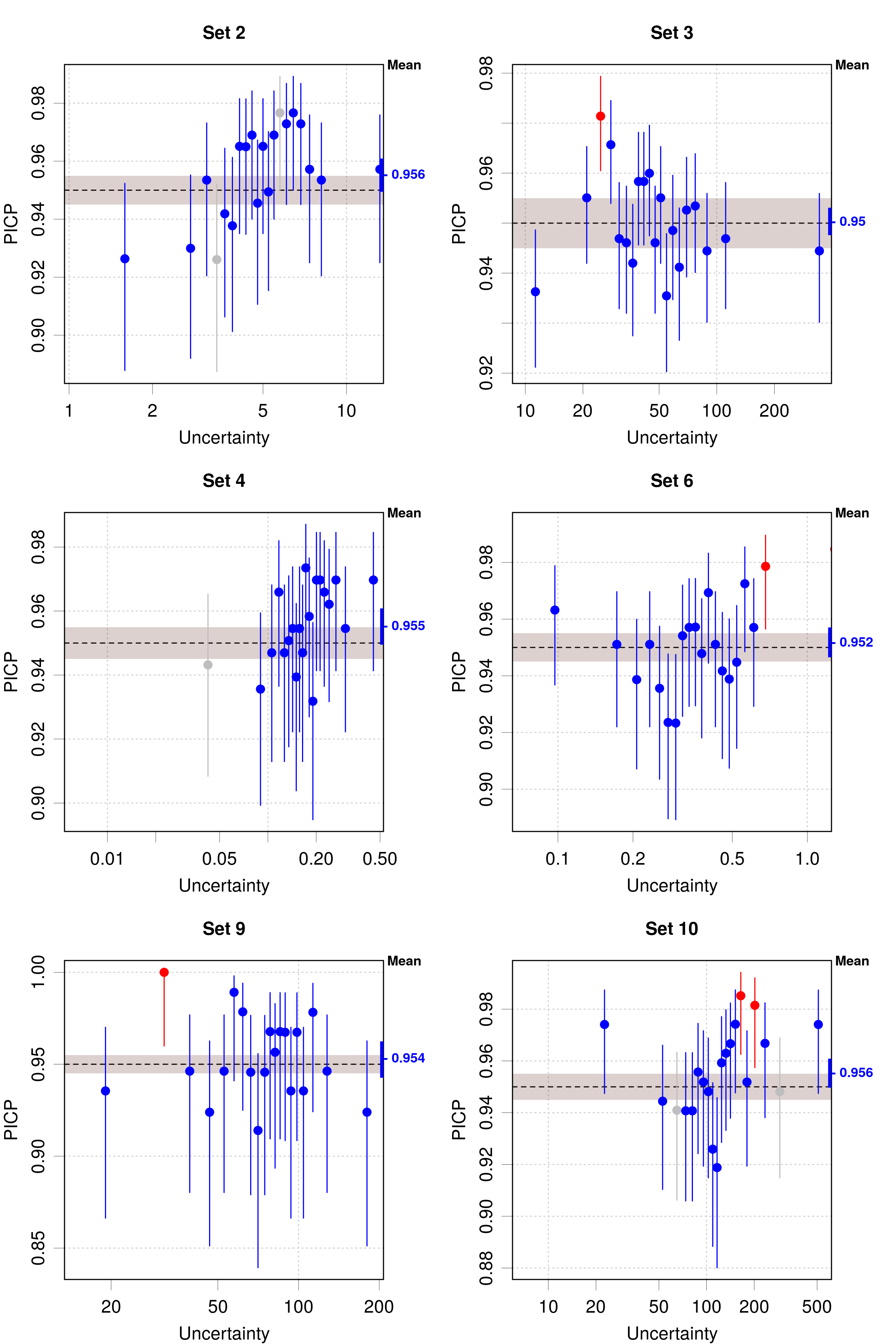}
\par\end{centering}
\caption{\label{fig:LCP-1}Local PICP (LCP) analysis using $N=20$ uncertainty-based
equal-size bins. The 95\,\% confidence intervals on the local PICP
values are reported as error bars. The average PICP value is reported
in the right margin. The gray area represents an admissible 0.005
deviation around the target value. Local values incompatible with
this admissible area are colored in red. Gray points cannot be reliably
tested.}
\end{figure}

\begin{figure}[t]
\noindent \begin{centering}
\includegraphics[width=0.9\textwidth]{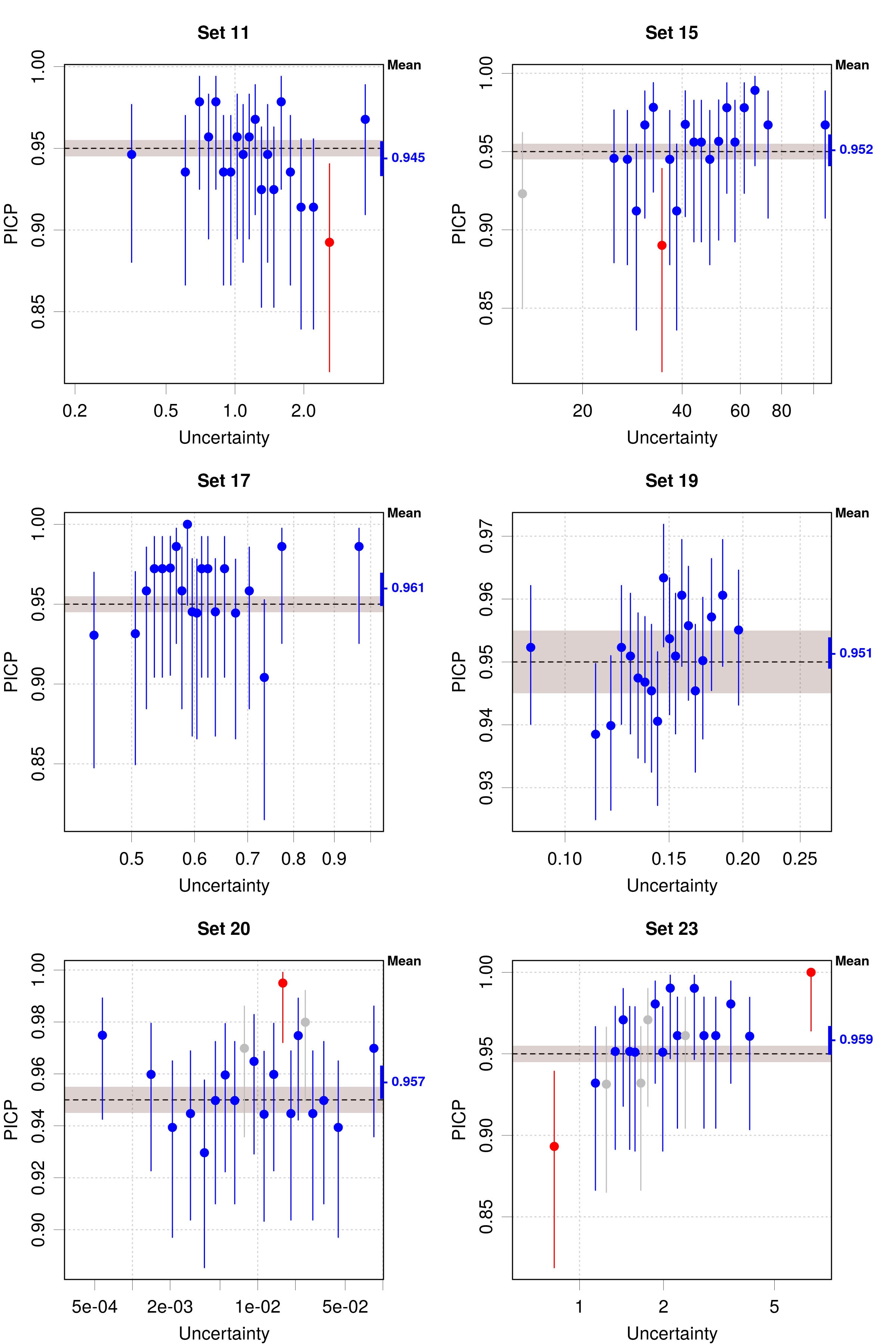}
\par\end{centering}
\caption{\label{fig:LCP-2}Fig.\,\ref{fig:LCP-1}, continued.}
\end{figure}

\begin{figure}[t]
\noindent \begin{centering}
\includegraphics[width=0.9\textwidth]{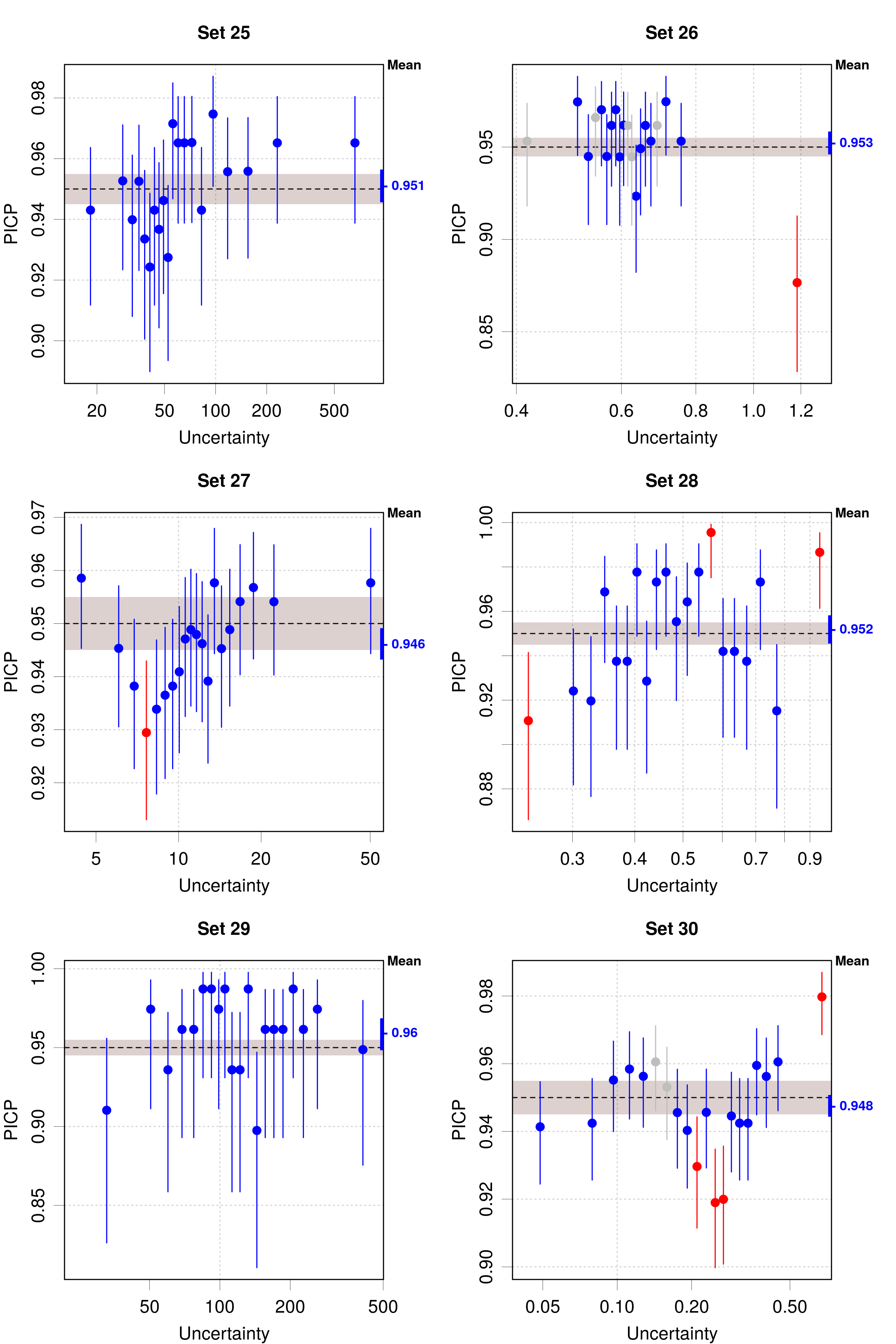}
\par\end{centering}
\caption{\label{fig:LCP-3}Fig.\,\ref{fig:LCP-1}, continued.}
\end{figure}

A local version of the PICP analysis, the LCP analysis\citep{Pernot2022b},
can be applied to the 18 datasets with validated PICP values to assess
their \emph{consistency}\citep{Pernot2023d}, using 20 equal-size
$u_{E}$-based bins (Figs.\,\ref{fig:LCP-1}-\ref{fig:LCP-3}). As
the post-hoc calibration used to design those datasets is based on
a polynomial correction of the uncertainties intended to correct for
major unsuitable trends in uncertainty space, on should expect a reasonable
consistency for most datasets.

The number of untestable bins per set is very small, varying from
0 to a maximum of 5 for Set 26. Similarly, very few bins are invalidated,
the most problematic case being Set 30 with 4 rejected bins. For this
set, uncertainty values between 0.2 and 0.3 seem to be consistently
underestimated. For Set 23, there seems to be a residual trend from
under- to over-estimation of $u_{E}$. Similar features are visible
for Sets 27 and 28, while Set 26 seems to suffer from a notable underestimation
of large uncertainties. 

\clearpage{}

\section{Conclusions}

Using an interval-based metric such as PICP to test prediction uncertainty
calibration offers more reliability and less computational burden
than using a variance-based metric such as ZMS. It was shown here
that interval information can simply be obtained from prediction uncertainty
due to the fact that \emph{z}-scores, even if often heavy-tailed,
have mostly scaled Student's $t_{s}(\nu)$ distributions. The proposed
estimation method of PICP at the 95\,\% level rests on the fact that
the $k_{95}$ enlargement factor for $t_{s}(\nu)$ is weakly dependent
on $\nu$ and can be fixed at 1.96 with negligible consequences, as
long as $\nu>3$. To avoid distribution fitting altogether, testable
datasets can be selected by a threshold on a robust skewness metric,
i.e. $\beta_{GM}(Z^{2})<0.85$. Unfortunately, this approach is not
applicable for other probability levels than 0.95.

Application to the 33 Jacobs \emph{et al.}'s datasets\citep{Jacobs2024}
shows that 10 sets have distribution properties (very heavy tails
or outliers) making them untestable for average calibration by the
PICP (or variance-based: ZMS, NLL, RCE...) metrics. To overcome this
limitation, an effort should be done to improve these distributions,
for instance by active learning\citep{Pernot2024}. Among the remaining
sets, 18 are validated for average calibration and 5 are not. For
the latter, the post-hoc polynomial transformation used by Jacobs
\emph{et al.}\citep{Jacobs2024} has not been fully efficient to ensure
calibration. Note however, that their PICP values do not exceed 0.97,
which might still be considered as acceptable. 

Consistency has been tested by the LCP analysis for the 18 calibrated
sets, showing that a large majority of them presents adequately uniform
local coverage in uncertainty space. Furthermore, this diagnostic
might help to improve the post-hoc calibration procedure for the few
datasets with local coverage problems.

\section*{Acknowledgments}

\noindent I warmly thank R. Jacobs for his help with the datasets.

\section*{Author Declarations}

\subsection*{Conflict of Interest}

The author has no conflicts to disclose.

\section*{Code and data availability\label{sec:Code-and-data}}

\noindent The code and data to reproduce the results of this article
are available at \url{https://github.com/ppernot/2024_PICP/releases/tag/v1.0}
and at Zenodo (\url{https://doi.org/10.5281/zenodo.13373267}). The
33 datasets of Jacobs \emph{et al.}\citep{Jacobs2024} are accessible
in a FigShare depository\citep{Morgan2024_FigShare}. 

\bibliographystyle{unsrturlPP}
\bibliography{NN}

\begin{thebibliography}{10}

\bibitem{Pernot2024}
P.~Pernot.
\newblock \href{http://dx.doi.org/10.48550/arXiv.2402.10043}{{Negative impact
  of heavy-tailed uncertainty and error distributions on the reliability of
  calibration statistics for machine learning regression tasks}}.
\newblock {\em arXiv:2402.10043}, February 2024.

\bibitem{Groeneveld1984}
R.~A. Groeneveld and G.~Meeden.
\newblock \href{http://dx.doi.org/10.2307/2987742}{Measuring skewness and
  kurtosis}.
\newblock {\em The Statistician}, 33:391--399, 1984.
\newblock URL: \url{http://www.jstor.org/stable/2987742}.

\bibitem{Bonato2011}
M.~Bonato.
\newblock \href{http://dx.doi.org/10.1016/j.frl.2010.12.001}{Robust estimation
  of skewness and kurtosis in distributions with infinite higher moments}.
\newblock {\em Financ Res Lett}, 8:77--87, 2011.

\bibitem{Pernot2021}
P.~Pernot and A.~Savin.
\newblock \href{http://dx.doi.org/10.1007/s00214-021-02725-0}{Using the {Gini}
  coefficient to characterize the shape of computational chemistry error
  distributions}.
\newblock {\em Theor. Chem. Acc.}, 140:24, 2021.

\bibitem{Jacobs2024}
R.~Jacobs, L.~E. Schultz, A.~Scourtas, K.~J. Schmidt, O.~Price-Skelly,
  W.~Engler, I.~Foster, B.~Blaiszik, P.~M. Voyles, and D.~Morgan.
\newblock \href{http://dx.doi.org/10.48550/arXiv.2406.15650}{{Machine Learning
  Materials Properties with Accurate Predictions, Uncertainty Estimates, Domain
  Guidance, and Persistent Online Accessibility}}.
\newblock {\em arXiv:2406.15650}, June 2024.

\bibitem{Pernot2022b}
P.~Pernot.
\newblock \href{http://dx.doi.org/10.1063/5.0109572}{Prediction uncertainty
  validation for computational chemists}.
\newblock {\em J. Chem. Phys.}, 157:144103, 2022.

\bibitem{Newcombe1998}
R.~G. Newcombe.
\newblock
  \href{http://dx.doi.org/10.1002/(SICI)1097-0258(19980430)17:8<857::AID-SIM777>3.0.CO;2-E}{Two-sided
  confidence intervals for the single proportion: comparison of seven methods}.
\newblock {\em Stat. Med.}, 17:857--872, 1998.

\bibitem{Pernot2022a}
P.~Pernot.
\newblock \href{http://dx.doi.org/10.1063/5.0084302}{The long road to
  calibrated prediction uncertainty in computational chemistry}.
\newblock {\em J. Chem. Phys.}, 156:114109, 2022.

\bibitem{Evans2000}
M.~Evans, N.~Hastings, and B.~Peacock.
\newblock {\em Statistical Distributions}.
\newblock Wiley-Interscience, 3rd edition, 2000.

\bibitem{Delignette2015}
M.~L. Delignette-Muller and C.~Dutang.
\newblock \href{http://dx.doi.org/10.18637/jss.v064.i04}{{fitdistrplus}: An {R}
  package for fitting distributions}.
\newblock {\em J Stat Softw}, 64(4):1--34, 2015.

\bibitem{Pernot2017}
P.~Pernot and F.~Cailliez.
\newblock \href{http://dx.doi.org/10.1002/aic.15781}{A critical review of
  statistical calibration/prediction models handling data inconsistency and
  model inadequacy}.
\newblock {\em AIChE J.}, 63:4642--4665, 2017.

\bibitem{Pernot2023d}
P.~Pernot.
\newblock \href{http://dx.doi.org/10.1063/5.0174943}{{Calibration in machine
  learning uncertainty quantification: Beyond consistency to target
  adaptivity}}.
\newblock {\em APL Mach. Learn.}, 1:046121, 2023.

\bibitem{Morgan2024_FigShare}
D.~Morgan and R.~Jacobs.
\newblock \href{http://dx.doi.org/10.6084/m9.figshare.26077015.v1}{{Machine
  Learning Materials Properties with Accurate Predictions, Uncertainty
  Estimates, Domain Guidance, and Persistent Online Accessibility - FigShare
  dataset}}.
\newblock 6 2024.

\end{thebibliography}

\clearpage{}

\appendix

\section{Distributions of $Z^{2}$ \label{sec:Distributions-of}}

\begin{figure}[t]
\noindent \begin{centering}
\includegraphics[width=0.9\textwidth]{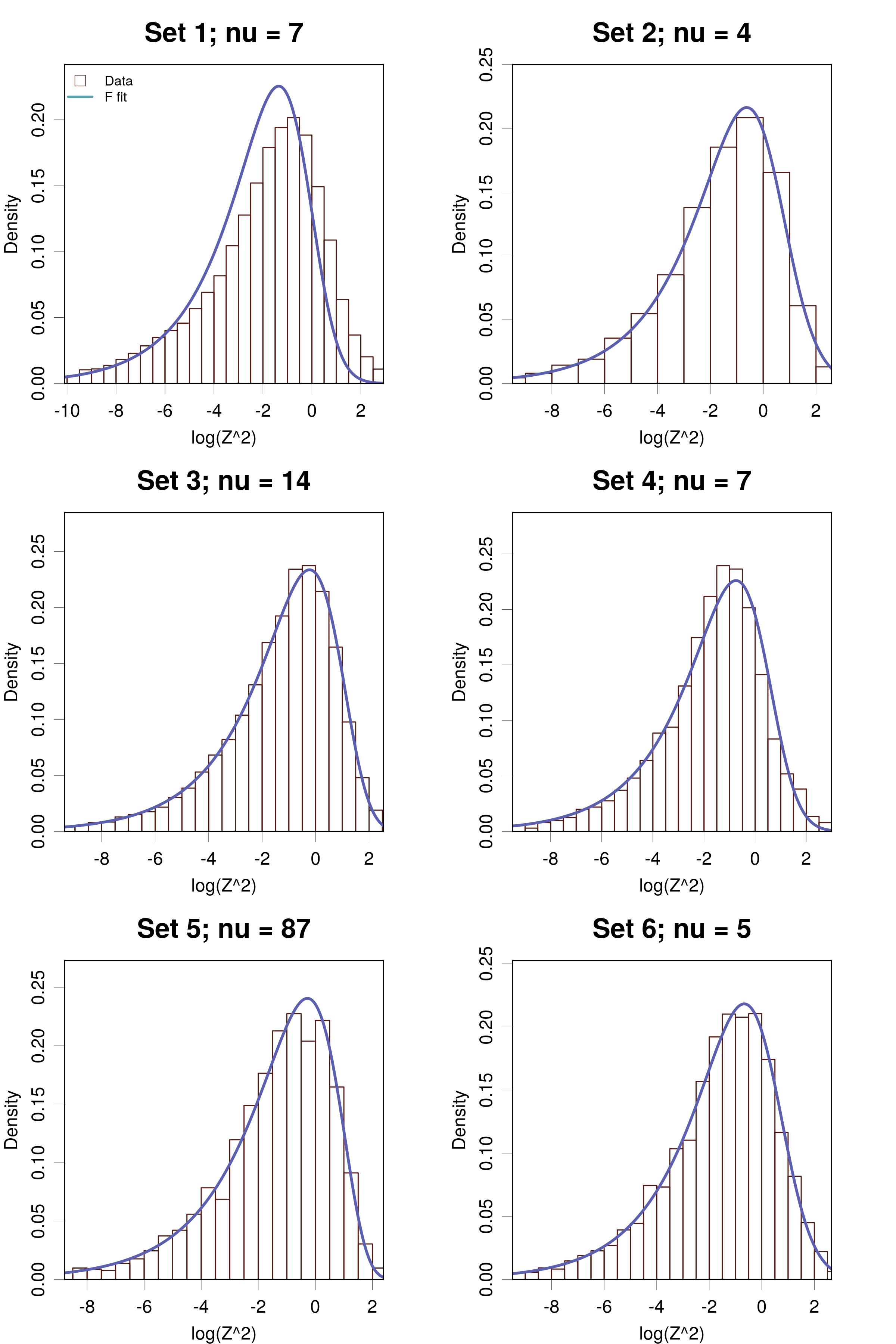}
\par\end{centering}
\caption{\label{fig:distZ2}Fit of the squared \emph{z}-scores (histogram)
by a scaled Fisher-Snedecor $F_{s}(1,\nu)$ distribution (blue line).
The rounded value of the best-fit shape parameter is reported in the
header of each plot.}
\end{figure}

\begin{figure}[t]
\noindent \begin{centering}
\includegraphics[width=0.9\textwidth]{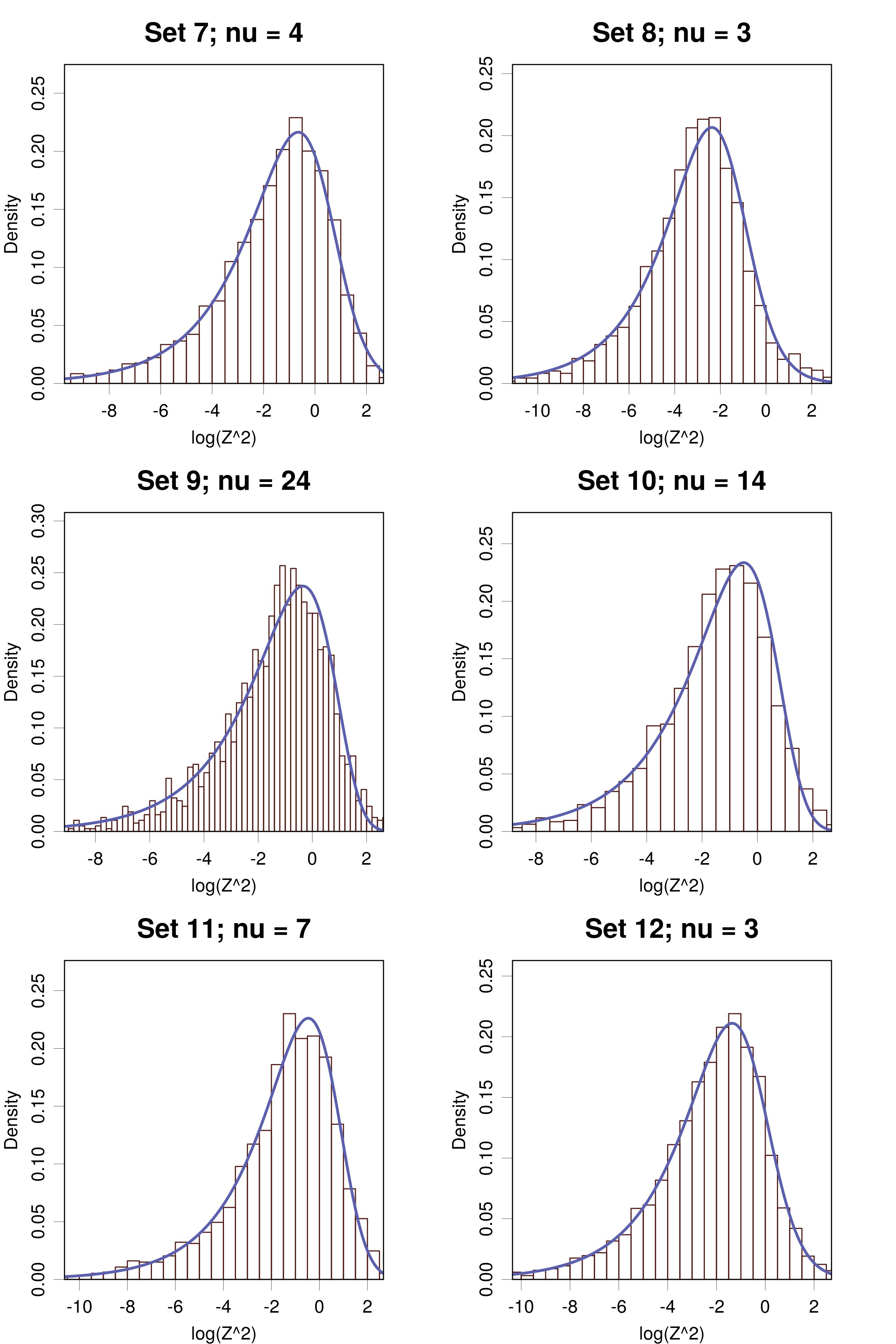}
\par\end{centering}
\caption{\label{fig:distZ2-1}Fig.\,\ref{fig:distZ2}, continued.}
\end{figure}

\begin{figure}[t]
\noindent \begin{centering}
\includegraphics[width=0.9\textwidth]{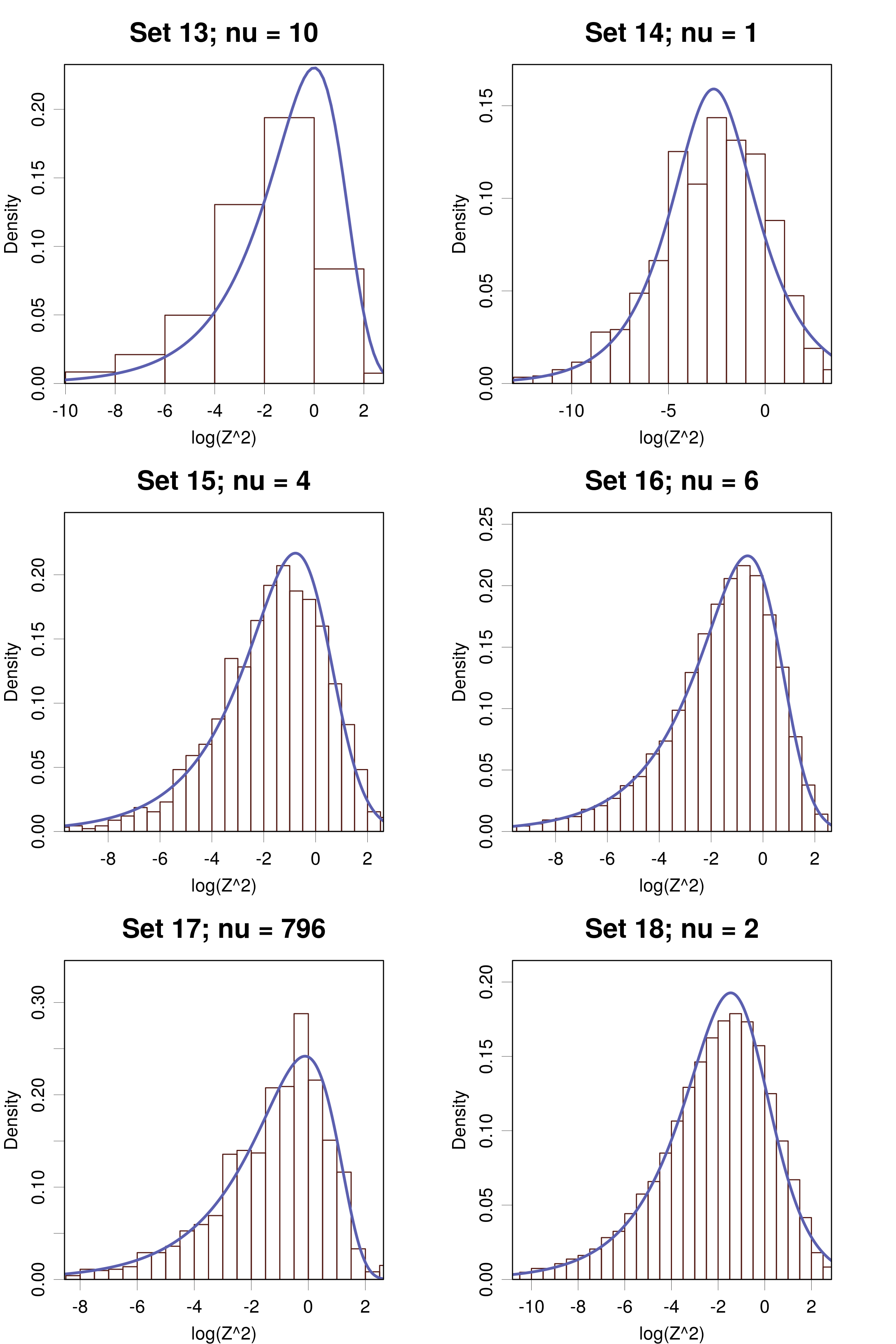}
\par\end{centering}
\caption{\label{fig:distZ2-2}Fig.\,\ref{fig:distZ2}, continued.}
\end{figure}

\begin{figure}[t]
\noindent \begin{centering}
\includegraphics[width=0.9\textwidth]{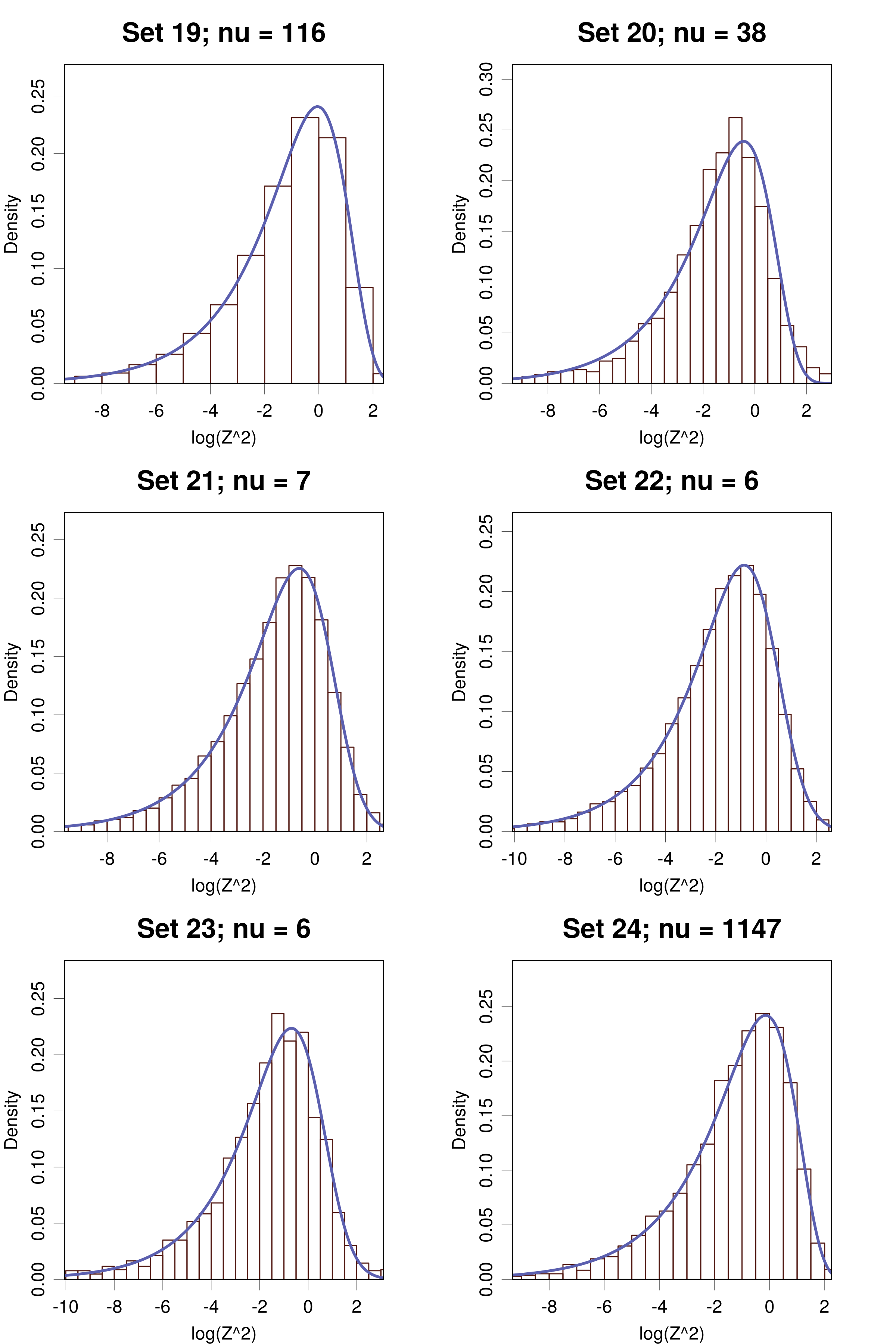}
\par\end{centering}
\caption{\label{fig:distZ2-3}Fig.\,\ref{fig:distZ2}, continued.}
\end{figure}

\begin{figure}[t]
\noindent \begin{centering}
\includegraphics[width=0.9\textwidth]{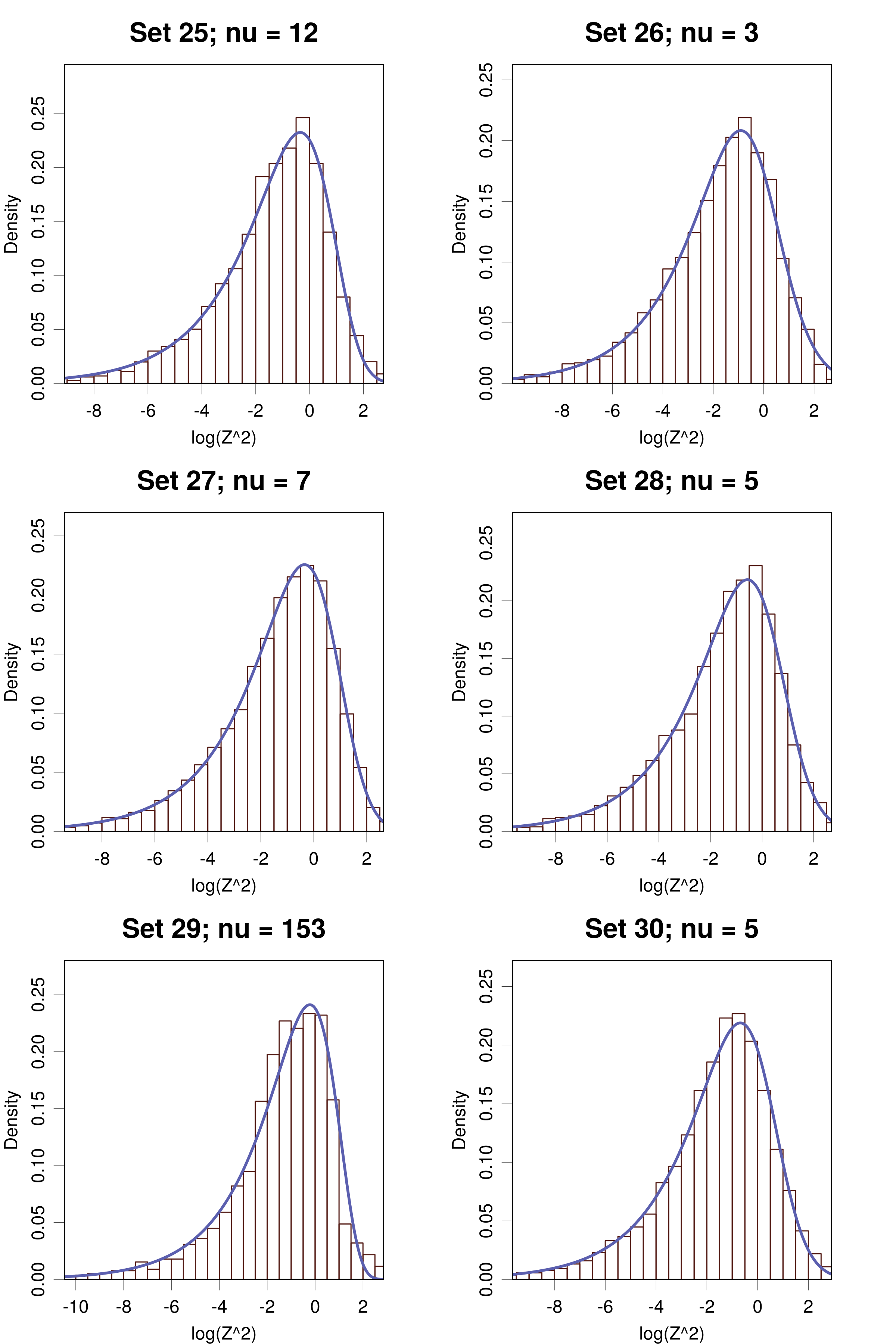}
\par\end{centering}
\caption{\label{fig:distZ2-4}Fig.\,\ref{fig:distZ2}, continued.}
\end{figure}

\begin{figure}[t]
\noindent \begin{centering}
\includegraphics[width=0.9\textwidth]{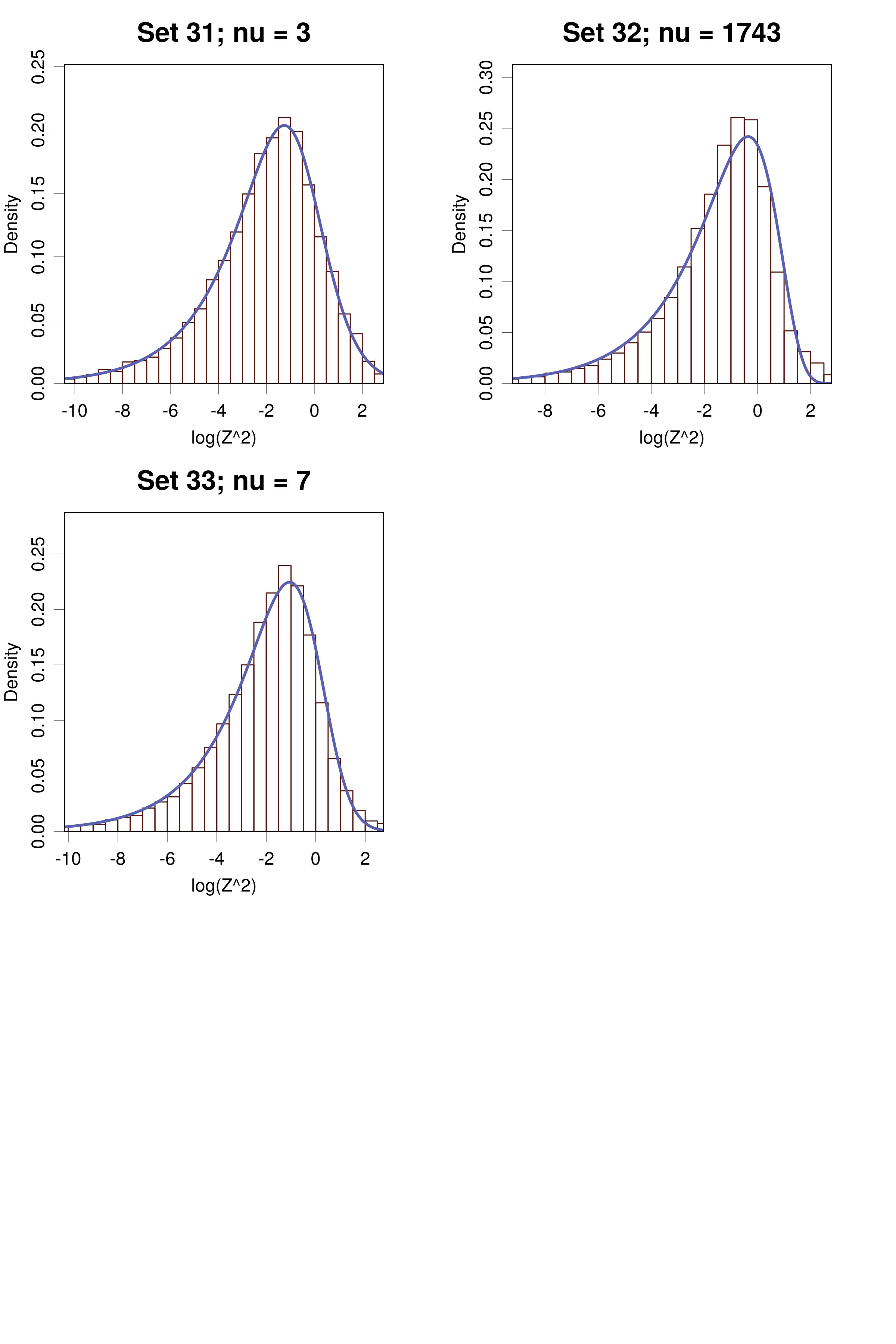}
\par\end{centering}
\caption{\label{fig:distZ2-5}Fig.\,\ref{fig:distZ2}, continued.}
\end{figure}

$Z^{2}$ datasets are fitted by a scaled $F$ distribution, using
a maximum goodness-of-fit estimation and the Kolmogorov-Smirnov distance
\citep{Delignette2015}. The results are reported in Figs.\,\ref{fig:distZ2}-\ref{fig:distZ2-5}. 


\end{document}